\documentclass[11pt,a4paper]{article}

\usepackage[hyperref]{acl2020}
\usepackage{times}
\usepackage{latexsym}
\usepackage{multirow}
\usepackage{graphicx}
\usepackage{subfig}
\usepackage{tikz}
\usetikzlibrary{shapes.geometric, arrows}

\usepackage{microtype}

\aclfinalcopy 

\title{DeepSentiPers: Novel Deep Learning Models Trained Over Proposed Augmented Persian Sentiment Corpus}

\author
{Javad PourMostafa Roshan Sharami, Parsa Abbasi Sarabestani, Seyed Abolghasem Mirroshandel\\
\normalsize{Department of Computer Engineering, University of Guilan, Rasht, Guilan, Iran}\\
\texttt{\{javad.pourmostafa,parsa.abbasi1996,mirroshandel\}@gmail.com}
}

\date{}

\begin{document}
\maketitle
\begin{abstract}
This paper focuses on how to extract opinions over each Persian sentence-level text. Deep learning models provided a new way to boost the quality of the output. However, these architectures need to feed on big annotated data as well as an accurate design. To best of our knowledge, we do not merely suffer from lack of well-annotated Persian sentiment corpus, but also a novel model to classify the Persian opinions in terms of both multiple and binary classification. So in this work, first we propose two novel deep learning architectures comprises of bidirectional LSTM and CNN. They are a part of a deep hierarchy designed precisely and also able to classify sentences in both cases. Second, we suggested three data augmentation techniques for the low-resources Persian sentiment corpus. Our comprehensive experiments on three baselines and two different neural word embedding methods show that our data augmentation methods and intended models successfully address the aims of the research.
\end{abstract}

\section{Introduction\label{intro}}
In recent years, a notable amount of subjective information has been accumulated on the internet. As an instance, online shopping markets where folks like to comment on products based on their satisfaction rate. In this case, extracting people's emotions from text has drawn much attention due to the broad range of its applications. On the one hand, discovering meaningful patterns from a large number of texts might be intrinsically complicated. Regarding this issue, sentiment analysis could boost the process of identifying patterns more useful than traditional file systems. Sentiment analysis, also called opinion mining, is the scope of research that analyzes people's opinions, sentiments, emotions towards entities such as products, services, organizations, and issues \cite{doi:10.2200/S00416ED1V01Y201204HLT016}. In fact, the text will be organized by the algorithm(s), and their outputs are described based on the rating score. They are also known as a polarity in terms of a set of subjective information like S, which commonly equals \{pos, negative, neutral\}. Hence, the task is known as a multi-class machine learning problem.

In sentiment analysis, supervised models are fed by a labeled dataset. Given that sentiment datasets are mostly English, it is needed to create non-English corpora. Preparing such a corpus takes time and requires native speakers to annotate the polarity. Alternatively, how well a suggested model predicts is another criterion. To address them, the research community has made many efforts to expand the number of data for further processing before applying the main text classification. Experts also evaluate their suggested models with the confusion matrix metrics. They all are chosen based on the purpose of the research. For instance, accuracy is one of them using when all the classes are equally important.

\begin{figure}[ht]
	\centering 
	\includegraphics[width=2.50in]{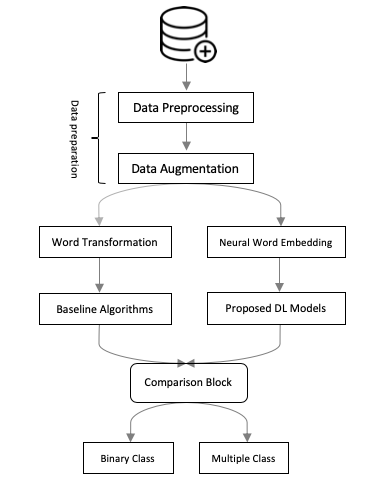} 
	\caption{An abstract perception of DeepSentiPers.} 
	\label{fig1}
\end{figure}

This paper consists of multiple modules so that each one of them interacts together to find the polarities on Persian (Farsi) sentences. We have specific criteria and certain limitations; However, the main aim of the research is defining Deep Learning (DL) architectures instead of other standard machine learning methods. So, an annotated dataset has been used to feed the models. Then, a few pre-processing steps have been applied to the entry. There is a matter which polarities should normally be within a reasonable percentage of each other. So, because of the unnormalized distribution in the dataset, the research proposes three data augmentation techniques giving a more balanced amount of classes. If we obtain dense and low-dimensional vectors, they may improve metrics by finding similar words. Hence, both pre-trained and online neural embedding have been utilized to determine semantic vectors. Finally, this study proposes two novel deep learning topologies that each one is proper for a specific output. They have covered both binary and multi-classes outputs. This abstract perception is depicted in Figure \ref{fig1}.

Accordingly, the outline of the paper is organized as follows. The related work about sentiment analysis on sentence-level is given in section 2. In section 3 and its sub-sections, data preparation is stated. Neural word embedding approaches are defined in section 4. In section 5, baseline and deep learning models are declared. Section 6 gives experimental results from two classification tasks. Eventually, Section 7 concludes the paper.

\section{Related Work}
Our work is closely related to sentence sentiment classification (SSC). In the literature, various research has been done to carry out novel sentiment models. Newly, most of them are based on deep learning algorithms, which can effectively increase the expected performance. 

For instance,  Kim et al. in \cite{Kim_2019} proposed a convolutional neural network (CNN) that consists of different layers to classify opinions over three datasets. Firstly, they initialized an embedding matrix, where words are located in space semantically and also completed through the training process. Secondly, the study used two convolutional layers that the first one stores local information, and the second one obtains features from contextual words according to the first layer. Afterward, the max-pooling has been used with an optional stride to fetch the most noticeable features. Finally, the probability value for each class with the leverage of a fully-connected (FC) layer and the softmax activation function has been calculated. Their suggested CNN model compared with traditional machine learning methods like Naive Bayes (NB), and Support Vector Machine (SVM). To sum up, the paper asserts that its offered model achieved 81\% and 68\% accuracies for binary and ternary classification, respectively.

Although the community has done some researches in terms of DNN's based approaches, there are a few Persian models that adequately handle its challenges. For example, a  \cite{7985281} used an unsupervised neural word embedding called Skip-gram \cite{2013arXiv1301.3781M} to represent words by their context window (C). The central part of the study is based on a Bidirectional Long Short Term Memory (BLSTM) and a CNN architecture. A `dropout' has been used to avoid overfitting as well. Then, the suggested CNN and BLSTM trained over a Persian dataset and assessed via a baseline named NBSVM-bi \cite{10.5555/2390665.2390688}. Finally, models reached 55.4\% and 53.2\% F-scores for binary sentiment classification, respectively. In this study, the collected data was unbalanced, which biased the learning process.

Among other few studies, Dashtipour et al. \cite{2018arXiv180805077D} specified a deep autoencoder and CNN for a novel Persian review dataset. After some common pre-processing steps, they converted words into vectors by the fasttext \cite{2016arXiv161203651J} as a library for text representation. Subsequently, an autoencoder has been employed to reduce dimensions and reconstruct the input. Their suggested CNN architecture has a total of 11 layers consists of convolution followed by max-pooling, fully-connected layers, and softmax activation function. Eventually, MLP has been compared with CNN and autoencoder. The paper claimed that the CNN architecture outperformed MLP and autoencoder with an accuracy of 82.6\% for binary classification.
\section{Data Preparation}

Choosing a proper SA corpus plays a vital role in classification of low resource languages. Inevitably the final prediction will be influenced by the specific domain of dataset.  Hence, the corpus used in this paper is called SentiPers~\cite{DBLP:journals/corr/abs-1801-07737}, which holds the customer reviews of an online shopping website named Digikala\footnote{https://digikala.com}. 
In SentiPers, each sentence of reviews is separated, and next will be annotated by four native speakers with features such as polarity, keywords, and targets. The polarity of each sentence selected from a set of emotion classes i.e. $E= \{-2, -1, 0, +1, +2\}$. See Table \ref{tab:set-ranked}, reflecting a set of ranked emotions.
\subsection{Parser Design}
For Preparing data, a particular parser \footnote{https://git.io/Jvh22} has been implemented regarding the XML structure of the given corpus. Then, all sentences and annotated polarities extracted subsequently. Finally, they have been accumulated into a single dataset file that contained $7415$ sentences in total. The number of sentences linked to each emotion is shown in Table~\ref{tab:dataset}. 

\subsection{Preprocessing}
\begin{figure}[ht]
	\centering 
	\includegraphics[width=1.60in]{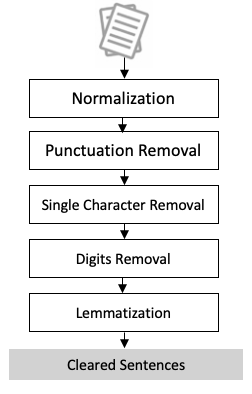} 
	\caption{Preprocessing steps} 
	\label{figpre}
\end{figure}

As it has been depicted in Figure \ref{figpre}, five different preprocessing steps were applied to all the sentences.
Firstly each sentence normalized using a Persian text preprocessing tool called Hazm. In this step, some white space misspelling was solved. In the three next steps, some punctuation and all the single characters and digits were removed from the sentence. Finally, as the last preprocessing step, each word in the text was replaced with their lemma.

\subsection{Binarization}
The given dataset has multiple emotions by default so that they can be operated directly in a multi-label classification task. However, for binary classifications, there must be just two groups. So, let $C$ denote the corpus and $s$ a subjective sentence, where $ s\in C$. Let $k$ also be the number of sentences in the given corpus and $C=\left\{ s_{1}, s_{2}, ..., s_{k} \right \}$. On the one hand, each $s$ linked to an emotion defined in set $E$. Merging sentences with either positive or negative class itself with no differences in rank concluding a ternary group in which classes are divided into negative, neutral, and positive categories.
Since there are three classes of emotion, the following approaches have been proposed:
\begin{enumerate}
\item NR: Removing neutral emotions 
\item NP: Considering neutral emotions as positive 
\item NN: Considering neutral emotions as negative 
\end{enumerate}

A reasonable approach is to ignoring neutral sentences and combining negative/positive classes into a single negative/positive class. Here, other methods (NP and NN) will be caused in rising the rate of unbalanced data as well. 
The number of sentences in each class after applying the selected approach is shown in Table~\ref{tab:binary-dataset}. 
It is important to announce that 75\% and 25\% of these sentences will be used as training data, and the testing data, respectively.
\begin{table}[]
\centering
\caption{Set of ranked emotions used in study.}
\label{tab:set-ranked}
\begin{tabular}{|c|c|}
\hline
\textbf{Ranked} & \textbf{Emotion Class} \\ \hline
-2 & Furious \\ \hline
-1 & Angry \\ \hline
0 & Neutral \\ \hline
+1 & Happy \\ \hline
+2 & Delighted \\ \hline
\end{tabular}
\end{table}

\begin{table}
\centering

\begin{tabular}{| l | c | c | c | c | c |}
\hline
\textbf{Ranked} & -2 & -1 & 0 & +1 & +2
\\\hline
\textbf{\#Sentences} & 40 & 697 & 3152 & 2184 & 1342
\\\hline
\end{tabular}
\caption{A number of sentences in each emotion.}\label{tab:dataset}
\end{table}

\begin{table}
\centering

\begin{tabular}{| l | c | c |}
\hline
\textbf{Polarity} & Negative & Positive
\\\hline
\textbf{\#Sentences} & 737 & 3526 
\\\hline
\end{tabular}
\caption{Binarized dataset statistics.}\label{tab:binary-dataset}
\end{table}

\subsection{Data Augmentation}
The performance of machine learning and deep learning often depends on the size and quality of training data, which is often tedious to collect \cite{2019arXiv190111196W}. So, for the obtained dataset which is small and somewhat imbalanced, three data augmentation (DA) methods have been proposed. To the best of our knowledge, they have not been used in the Persian language so far. The selected DA method will be introduced in the result section, subsequently.
\subsubsection{Extra Data Augmentation (Balanced)}
\begin{figure}[ht]
	\centering 
    \subfloat[original dataset]{{\includegraphics[width=3.35cm]{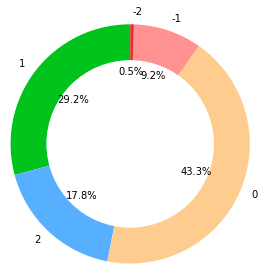} }}%
    \qquad
    \subfloat[balanced dataset]{{\includegraphics[width=3.35cm]{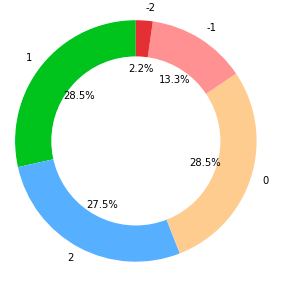} }}%
	\caption{Dataset distribution.} 
	\label{dataset-distribution}
\end{figure}

 Extra DA aims to modify the dataset by using another version of the SentiPers package including additional data. Consequently, the number of sentences in the poor classes is increased by applying the extra data and a few numbers of sentences in the rich classes are decreased. The results are shown in Fig~\ref{dataset-distribution}.

\subsubsection{Translation Data Augmentation}
This approach focuses on increasing data through data noising technique \cite{2017arXiv170302573X}. Viewing that in this case there are common and different methods for working on images and sounds. But text DA will be faced with constraints like grammatical structure, vocabularies, and pragmatic meanings.
Considering we are seeking for sentiments, the order, as well as the structural role of words, would seldom affect the process. So, it is required to focus on those words used regularly in a specific polarity. Finally, the sentence sentiment can be deducted by translated words evoking from dictionaries like Google Translator (GT).

\begin{figure}[ht]
	\centering 
	\includegraphics[width=2.50in]{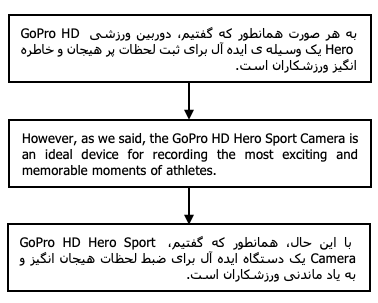} 
	\caption{Translation Data Augmentation} 
	\label{fig2}
\end{figure}

According to this hypothesis, every single sentence in the training data is translated into a middle-language (here English) by the GT. Then, it will be re-translated to its initial Persian form \cite{2017arXiv170500440F}. In the proposed process, although some words change to their synonyms, the word position switch and so many unpredictable changes occur, the polarity and sentiment of the entire sentence will be intact. Applying the suggested method double the given data set. An example of translation DA has been shown in Figure \ref{fig2}.

\subsubsection{Synonym Data Augmentation}
Contrary to the last approach, this method aims to substitute only a few words with their synonyms. On the other hand, the GT often generates synonyms as well as a list of re-translation to the initial language (Persian). Considering this idea, firstly, the floor of 20\% of each sentence is selected randomly for substitution. It is worth mentioning that some text pre-processing steps such as space to ZWNJ\footnote{Zero-Width Non-Joiner} conversion and punctuation removal already took place. Secondly, every selected word is translated and generates a list of synonyms subsequently. Then, again a random word is chosen for its synonym list. 

\begin{figure}[ht]
	\centering 
	\includegraphics[width=2.40in]{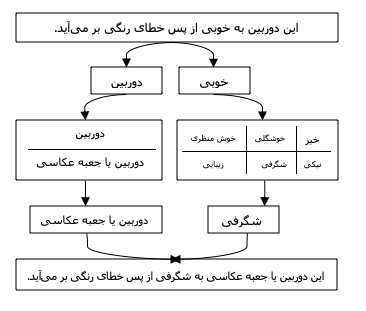} 
	\caption{Synonym Data Augmentation} 
	\label{fig3}
\end{figure}

The method output might be different per every single runtime in that using two random selection strategies. As a result, this method like the translation DA will inevitably double data. In this case, Figure \ref{fig3} illustrates the cycle of synonym DA. 

\section{Neural Word Embedding}
Working with text data requires a method to convert them to numbers since each model has a mathematical base. Despite sparse vectorization that often uses in traditional machine learning algorithms, we benefit from neural word embedding in deep learning models where each word is mapped to a low-dimension vector called word feature \cite{turian-etal-2010-word}. Defining words in this representation makes finding similar words directly. So we can also use this advantage for opinion mining. 

In a nutshell, each sentence is tokenized, the next words will be vectorized. Consequently, the more similar vectors met among sentences, the more similarity they would have. It is worth mentioning that neural word embedding is not only a synonym detector but also a method for finding words from the same family (e.g. cat, dog). These vectors are learned in two possible ways as follows.

\subsection{Online Embedding Layer}
\begin{figure}[ht]
	\centering 
	\includegraphics[width=2.50in]{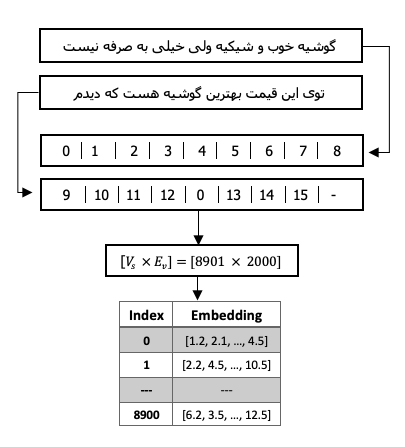} 
	\caption{Online Embedding Layer} 
	\label{fig4}
\end{figure}

This method relies on the available dataset and will be operated in the neural learning process. In fact, the output vectors are not computed from the input using any mathematical operation. Thus, each word in sentences is encoded by a unique integer number as they appeared. In this case, let $V_s$ denote the number of words in the vocabulary set and $E_v$ shows the dimension of embedding vectors. Then, once the neural network has been trained, we expect an output embedding vector of size as follows. In the meantime, Figure \ref{fig4} illustrates how online embedding layer works for two sentences.
\[[V_s \times E_v]\]

\subsection{Pre-trained Word Embedding}
Contrary to the online neural embedding, this technique utilizes pre-trained embedding vectors. Until writing this research, several pre-trained neural vectors were made by universities as well as companies. Among them, the appropriate choice for the Persian language would be a FastText library provided by Facebook \cite{bojanowski2017enriching} \cite{joulin2016fasttext}.
Viewing that the data used in FastText collected from Persian Wikipedia \footnote{https://fa.wikipedia.org/} so that each word is mapped to a 300-dimension vector.

\section{Models}
\subsection{Baseline Models}
Generally, the new model is assessed by well-known baseline patterns. Viewing that sentiment analysis known as a common text classification problem. So, three machine-learning algorithms including naive Bayes classifier \cite{10.5555/645526.657278}, stochastic gradient descent \cite{Prasetijo2017HoaxDS} and support vector machine \cite{lili} have been selected. Finally, the proposed deep learning outputs are compared with baselines in terms of evaluation metrics.

\subsection{Deep Learning Models}
Recently, deep learning has been used in many fields of natural language processing \cite{10.5555/1953048.2078186} where they do not need any manual feature extraction as well as human resources \cite{7748849}. They can also classify datasets in both supervised and unsupervised methods. 
Among common DL architectures, since Long-Short-Term Memory (LSTM) benefit from an internal memory unit, it can hold recent data. On the other hand, Convolutional Neural Networks (CNN) can also extract local features through the context window. Considering the existing advantages, we propose the Persian sentiment analysis models powered by LSTM and CNN.

In more detail, considering that texts are made of embedded words, in the first layer of the suggested models, the number of neurons equals to the maximum length of sentences tokenized by word. In our dataset, the longest text contains 257 words, so all models have 257 neurons in their first layer.
The next layer is called the embedding layer where every single word is embedded in the multi-dimensional vector. It is important to note the size of this layer in models trained by FastText embedding has been fixed in 300-dimension previously. However, the suggested models utilizing Keras word embedding has an output with 2000 neurons.

\subsubsection{BLSTM Architecture}
\begin{figure}[ht]
	\centering 
	\includegraphics[width=3.15in]{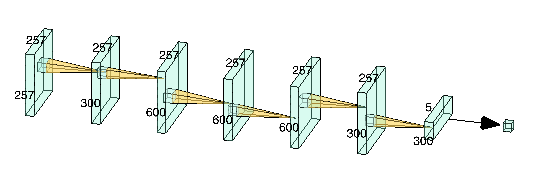} 
	\caption{A Schematic Diagram of Suggested Persian Sentiment Model Using BLSTM Architecture, generated via \cite{LeNail2019}} 
	\label{blstm}
\end{figure}

The first suggested deep learning architecture relies on recurrent neural networks called Bidirectional-LSTM. Apart from two initial layers (input and embedding) mentioned in the previous section, this Persian sentiment model consists of six more extra layers as follows: 
\begin{enumerate}
    \item \textbf{Bidirectional CuDNNLSTM} which is fast LSTM implementation backed by NVIDIA CuDNN (run on GPU with TensorFlow backend).
    \item \textbf{Global max-pooling 1D} which takes two-dimensional tensor of input size and input channels and computes the maximum of all the values for each of the input channels. It is important to note that stride equals one.
    \item \textbf{Dropout} \cite{JMLR:v15:srivastava14a} regularization to prevent overfitting by randomly 10 percent drop in the input.
    \item \textbf{Dense} which is a regular deeply connected neural layer. Let \textit{y}, \textit{w} and \textit{b} denote the output, weight and bias, respectively. Then, the following equation does on the input (\textit{x}): 
\[y = \sum_{i=1}^{n} X_{i}W_{i}+b\]
Once the output has been computed, the ReLU \cite{pmlr-v15-glorot11a} is used as an activation function . Thus, the final output would be: 
 \[y = max(0, y)\]It is also worth mentioning that \textit{n} represents the total number of neurons and equals 600 in the model.
    \item A new \textbf{Dropout} layer, however with the same configuration as mentioned before.
    \item A new \textbf{Dense} layer with the different activation function called Softmax \cite{2018arXiv181103378N}. Given that the proposed models are able to classify polarity in both binary and multi-label classes, the output of last dense layer is either five or two.
\end{enumerate}
As an instance, the illustration of the suggested BLSTM model implemented by FastText word-embedding has been shown in Figure \ref{blstm}.
\subsubsection{CNN Architecture}
The second suggested deep learning stack benefits from CNN. This architecture usually uses in the scope of image processing. However a survey has shown by \cite{kim-2014-convolutional} that it can be also efficient for text classification. So, the proposed model is comprised of nine more extra layers including CNN as follows:
\begin{enumerate}
    \item \textbf{Conv 1D} which creates a convolution kernel that is convolved with the layer input over a single dimension to produce a tensor of outputs. In this layer, the kernel size equals 4 and filters are 64. It is also important to note that the ReLU has been used as an activation function once the layer computed the output.
    \item \textbf{Max pooling 1D} which is different from its global form. In more detail, it takes a pool length, but global max pooling does not. Given that stride is one and pool size equals 2, then the maximum value of each slide with length 2 will be selected.
    \item \textbf{Conv 1D} which is similar to the previous convolutional layer. However, its kernel size is 8 which means that 8 consecutive elements are being considered and they finally produce a single value.
    \item \textbf{Max pooling 1D} which has been described beforehand with no differences in parameters. 
    \item \textbf{Conv 1D} which is the same as others. But, its kernel size has been changed to 16.
    \item \textbf{Global max-pooling 1D} which functions as it has been described in the LSTM architecture.
    \item \textbf{Dropout} which drops out nodes during training time by 10 percent. It is worth mentioning that the higher the dropout rate is, the more generalized model will be made. So, the percentage has been selected based on a trade-off.
    \item \textbf{Dense} which uses the Sigmoid as an activation function. Let \textit{y} denote the output of the model before using activation function. Then, \textit{f(x)} represent: \[f(x) = \frac{1}{1+x^{-y}}\]
    \item \textbf{Dense} which returns either a binary classification or a classification of emotions -2 to +2.
\end{enumerate}

\section{Results}
Given that the study comprises of various stages in terms of data augmentation, word-embedding, and deep learning architectures, the metrics should be computed in all forms of existing combinations. It is important to note that, the suggested NR method has been selected to report the binary results. 

As it has been stated in Section \ref{intro}, our models are able to classify both binary and multi-label classes. So, the consequences have been reported in two separate sections regarding their connected baselines. It is also worth mentioning that among suggested data augmentation techniques, the translation approach has been selected for both binary and multiple results. However, a full report of experiments is available on the project's repository for further studies.

\subsection{Metric}
We reckon \textit{F1-score} is a pertinent metric to evaluate the proposed models since the class distribution is imbalanced. Let P denote precision and R represent recall, then f1-score is calculated using a weighted averaging method that relies on the number of true labels of each class by the following equation. 
\[F1-Score = 2\times \frac{P\times R}{P+R}\]

\subsection{Binary Results}
In this type of result, the output is restricted into two different opinions (positive or negative). Among our provided models, the prominence of  B-LSTM embedded by Keras is striking. As it has been illustrated in Table \ref{tab:binary-result}, the sentiment model implemented by translation data augmentation has the highest rate with 91.986\% f1-score compared to its baseline with only 91.312\%.

\begin{table*}[]
\centering
\begin{tabular}{c|c|c|c|l|c|l|l|c|c|c}
\hline
\multirow{3}{*}{\textbf{Dataset}} & \multicolumn{4}{c|}{\multirow{2}{*}{\textbf{Baseline Models}}}    & \multicolumn{6}{c}{\textbf{Deep Learning}}                                                      \\ \cline{6-11} 
                                  & \multicolumn{4}{c|}{}                                             & \multicolumn{4}{c|}{\textbf{Keras Embedding}} & \multicolumn{2}{c}{\textbf{FastText Embedding}} \\ \cline{2-11} 
                                  & \textbf{NB}               &\textbf{SVM}               & \multicolumn{2}{c|}{\textbf{SGD}}   & \multicolumn{3}{c|}{\textbf{B-LSTM}}   & \textbf{CNN}           & \textbf{B-LSTM}                  & \textbf{CNN}                    \\ \hline
\textbf{Original}                          & 74.39            & 83.81            & \multicolumn{2}{c|}{74.34} & \multicolumn{3}{c|}{85.59}   & 84.78        & 81.152                  & 79.90                 \\ \hline
\textbf{Balanced}                          & 75.45            & 91.77            & \multicolumn{2}{c|}{74.34} & \multicolumn{3}{c|}{91.65}   & 90.502        & 89.268                  & 87.31                 \\ \hline
\textbf{Translation}                       & 76.67            & \textbf{91.31}            & \multicolumn{2}{c|}{74.39} & \multicolumn{3}{c|}{\textbf{91.98}}   & 91.906        & 90.592                  & 88.06                 \\ \hline
\end{tabular}
\caption{Binary classification results based on weighted average of the F1 scores.}\label{tab:binary-result}
\end{table*}

\subsection{Multi-class Results}
Just to remind you, initially the dataset has been labeled by five emotions ranged from -2 to +2. As Table \ref{tab:multiclass-result} indicates, again the SVM model embedded by extra data augmentation is our flagships among baselines. 68.344\% f1-score has been recorded for its multi-classification rate. On one hand, the top f1-score goes to the suggested B-LSTM model using translation DA and embedded by FastText with 69.33\%. It is higher than its relevant machine learning baseline which is 67.62\%.

\begin{table*}[]
\centering
\begin{tabular}{c|c|c|c|l|c|l|l|c|c|c}
\hline
\multirow{3}{*}{\textbf{Dataset}} & \multicolumn{4}{c|}{\multirow{2}{*}{\textbf{Baseline Models}}}                 & \multicolumn{6}{c}{\textbf{Deep Learning}}                                                                  \\ \cline{6-11} 
                                  & \multicolumn{4}{c|}{}                                                          & \multicolumn{4}{c|}{\textbf{Keras Embedding}}             & \multicolumn{2}{c}{\textbf{FastText Embedding}} \\ \cline{2-11} 
                                  & \textbf{NB}         & \textbf{SVM}         & \multicolumn{2}{c|}{\textbf{SGD}} & \multicolumn{3}{c|}{\textbf{B-LSTM}} & \textbf{CNN}       & \textbf{B-LSTM}          & \textbf{CNN}          \\ \hline
\textbf{Original}                 & 40.36               & 62.47               & \multicolumn{2}{c|}{54.65}        & \multicolumn{3}{c|}{64.98}          & 63.006             & 64.72                   & 63.364                \\ \hline
\textbf{Balanced}                 & 56.04               & 68.34               & \multicolumn{2}{c|}{65.34}       & \multicolumn{3}{c|}{66.02}          & 66.366             & 68.85                   & 65.806                \\ \hline
\textbf{Translation}              & 56.65               & \textbf{67.62}                & \multicolumn{2}{c|}{62.04}       & \multicolumn{3}{c|}{66.50}          & 66.654             & \textbf{69.33}                   & 66.306                \\ \hline
\end{tabular}
\caption{Multi-class classification results based on weighted average of the F1 scores.}\label{tab:multiclass-result}
\end{table*}

\section{Conclusion}
It is an idea that deep learning models require a large amount of data. So, three data augmentation techniques have been suggested. Among them, translation DA got the highest rank. In other words, our research simply supported the initial idea and provided a solution for a lack of existing annotated corpus in the field of Persian sentiment analysis. Besides, It is expected that the multiple classifications can likely have an outstanding growth in bigger training data.

Because both binary and multiple classifications are varied in terms of word embedding, our research also shows that we can not be cognizant of using an absolute word embedding method and vouch for its superiority. In fact, each embedding technique works in a specific training domain, which provides this opportunity to make an intelligent and changeable embedding method. 

The main achievement is our proposed deep learning stacks which both obtained appropriate f1-scores. But, B-LSTM works better for sentiment classification due to its internal memory units that are able to handle long dependencies. The research also confirm this idea that CNN's backbone is appropriate for the scope of image processing rather than text classification. So, more future researches can be conducted in bidirectional LSTM regarding its connectivity with sentiment analysis.

\bibliography{acl2020}
\bibliographystyle{acl_natbib}

\appendix

\end{document}